\documentclass{opt2024} 

\usepackage{booktabs}

\title[Hierarchical Simplicity Bias of Neural Networks]{Hierarchical Simplicity Bias of Neural Networks}

\optauthor{%
\Name{Zhehang Du} \Email{duz@wharton.upenn.edu}\\
\addr The Wharton School, University of Pennsylvania}

\begin{document}
\maketitle

\begin{abstract}
Neural networks often exhibit simplicity bias, favoring simpler features over more complex ones, even when both are equally predictive. We introduce a novel method called \emph{imbalanced label coupling} to explore and extend this simplicity bias across multiple hierarchical levels. Our approach demonstrates that trained networks sequentially consider features of increasing complexity based on their correlation with labels in the training set, regardless of their actual predictive power. For example, in CIFAR-10, simple spurious features can cause misclassifications where most cats are predicted as dogs and most trucks as automobiles. We empirically show that last-layer retraining with target data distribution \citep{kirichenko2022last} is insufficient to fully recover core features when spurious features perfectly correlate with target labels in our synthetic datasets. Our findings deepen the understanding of the implicit biases inherent in neural networks.
\end{abstract}


\section{Introduction}

Neural networks (NNs) have demonstrated remarkable capabilities in learning and generalizing from data, even in over-parameterized settings, a phenomenon known as double descent \citep{nakkiran2021deep}. Despite their adeptness, neural networks can exhibit vulnerability when faced with real-world challenges like distribution shifts \citep{hendrycks2021many} and adversarial attacks \citep{szegedy2013intriguing,madry2017towards}. One underlying reason for these vulnerabilities is the \emph{simplicity bias} \citep{valle2018deep, kalimeris2019sgd, shah2020pitfalls, pezeshki2021gradient}, where gradient descent tends to favor learning simple, strongly correlated features over more complex but robust ones. This preference for simplicity might seem reasonable, as this inductive bias naturally reflects the properties of real-world data and leads to generalization \citep{teney2022evading}. However, it can lead to models that struggle when subjected to distributional shifts and adversarial attacks.

\paragraph{Background.} Simplicity bias is the tendency of neural networks to learn simple functions, potentially ignoring more complex but equally or more predictive features \citep{valle2018deep,shah2020pitfalls}. This phenomenon is illustrated in the work by \citet{shah2020pitfalls}, where neural networks were shown to preferentially learn from simpler, more salient features at the expense of more complex but equally predictive ones. For example, in their MNIST-CIFAR dataset, images in class $-1$ are concatenations of MNIST digit zero and CIFAR-10 automobiles, while images in class $1$ are concatenations of MNIST digit one and CIFAR-10 trucks. It turns out that the trained network only depends on the MNIST digit for classification. Another example is found in shortcut learning \citep{geirhos2020shortcut}, where the neural network focuses on object location rather than object type. 

\paragraph{Motivation.} Understanding how neural networks learn in the presence of spurious features is critical, especially when these features introduce false associations with target labels, degrading model performance \citep{sagawa2019distributionally}. Previous studies have focused on mitigating these vulnerabilities, but the hierarchical nature of simplicity bias across multiple feature complexities remains underexplored.

\paragraph{Contributions.} In this study, we introduce \emph{imbalanced label coupling} to extend the concept of simplicity bias to multiple hierarchical levels. Our key contributions are: (a) We demonstrate that neural networks exhibit a \emph{hierarchical simplicity bias}, making predictions by sequentially considering features of increasing complexity, akin to decision trees. (b) We provide empirical evidence using synthetic datasets designed to highlight hierarchical decision-making based on feature complexity. (c) We show that last-layer retraining with target data distribution \citep{kirichenko2022last} is insufficient to fully recover core features when spurious features perfectly correlate with target labels, highlighting limitations in addressing hierarchical simplicity bias.

\section{Formulation of Hierarchical Simplicity Bias}
\label{sec:def}

\paragraph{Notation.} We formally define the notations used in our formulation: $\mathbf{x} = (\mathbf{x}_s, \mathbf{x}_c)$ represents the input data, where $\mathbf{x}_s$ denotes simple features (e.g., MNIST digits or patches), and $\mathbf{x}_c$ denotes complex features (e.g., CIFAR-10 images). The symbol $y \in \mathcal{Y}$ indicates the true label associated with $\mathbf{x}$, while $\hat{y} \in \mathcal{Y}$ represents the predicted label by the neural network. The function $f: \mathcal{X}_s \times \mathcal{X}_c \rightarrow \mathcal{Y}$ is the neural network's prediction function. Furthermore, $\mathcal{Y}_{\mathbf{x}_s} \subseteq \mathcal{Y}$ specifies the subset of labels associated with the simple feature $\mathbf{x}_s$ due to imbalanced label coupling.

\subsection{Hierarchical Decision Process}

We introduce an idealized hierarchical decision process where neural networks perform predictions akin to a decision tree, sequentially considering features of increasing complexity.

\paragraph{Training: Imbalanced Label Coupling.} We construct the training set by coupling classes from two different datasets in an imbalanced manner. For each class from the coarse dataset (e.g., MNIST digits), we concatenate it with multiple classes from the fine dataset (e.g., CIFAR-10 images) to create training examples. The labels are assigned \textbf{solely based on the fine dataset}, while the coarse dataset introduces spurious correlations. 

Depending on which dataset serves as the fine dataset, we consider two scenarios. In Scenario A, the fine dataset corresponds to the complex features \( \mathbf{x}_c \), and the labels are assigned based on \( \mathbf{x}_c \). The simple features \( \mathbf{x}_s \) (from the coarse dataset) introduce spurious correlations. The true label \( y \) is determined by the complex features \( \mathbf{x}_c \), i.e., \( y = y(\mathbf{x}_c) \). Likewise, in scenario B, the fine dataset is simple features. For example, in scenario A, if MNIST digit 1 is paired with CIFAR-10 classes automobile and cat, then any image containing digit 1 can be labeled as either automobile or cat based on the CIFAR-10 image it is paired with.

\paragraph{Testing: Hierarchical Decision Process.}
The test set is created by concatenating the image channels from all selected classes in each dataset, \textbf{without any coupling constraints}. During testing, the neural network's prediction $\hat{y}$ follows a hierarchical decision process:
\begin{itemize}
    \item \textbf{Scenario A:} The coarse features \( \mathbf{x}_s \) are correlated with subsets of labels. For each value of \( \mathbf{x}_s \), there is an associated subset \( \mathcal{Y}_{\mathbf{x}_s} \subseteq \mathcal{Y} \). The prediction function is:
    \begin{equation}
    \hat{y} = f(\mathbf{x}_s, \mathbf{x}_c) = f_{\mathbf{x}_s}(\mathbf{x}_c) = \mathop{\arg\max}_{y' \in \mathcal{Y}_{\mathbf{x}_s}} p(y' \mid \mathbf{x}_c),
    \label{eq:scenario_A_prediction}
    \end{equation}
    where \( f_{\mathbf{x}_s} \) denotes the decision function conditioned on \( \mathbf{x}_s \), mapping \( \mathbf{x}_c \) to a label within \( \mathcal{Y}_{\mathbf{x}_s} \). First, the network uses the coarse features to narrow down the possible labels to a subset. Then, within this subset, the network uses the fine features to predict the final label \( \hat{y} \).

    \item \textbf{Scenario B:} The labels are assigned based on the simple features $\mathbf{x}_s$, which are fully predictive. Due to simplicity bias, the network only depends on the simple features \( \mathbf{x}_s \) for prediction. The prediction function then simplifies to:
    \begin{equation}
    \hat{y} = f(\mathbf{x}_s) = \mathop{\arg\max}_{y' \in \mathcal{Y}} p(y' \mid \mathbf{x}_s).
    \label{eq:scenario_B_prediction}
    \end{equation}
\end{itemize}

In both scenarios, the decision-making prioritizes simple features, i.e., making decisions according to the ascending complexity of features.

\subsection{Quantitative Measures of Hierarchical Simplicity Bias}

Given that neural networks trained with gradient descent by empirical risk minimization may not perfectly align with the hierarchical prediction, we introduce quantitative measures to more accurately capture the hierarchical simplicity bias observed in the confusion matrix. We define separate measures for each scenario to account for the differences in how the hierarchical simplicity bias manifests.

\begin{itemize}
\item \textbf{Scenario A.} For each true label \( y \) and associated simple feature \( \mathbf{x}_s \), we define the \textbf{Hierarchical Classification Accuracy (HCA)} as:
\begin{equation}
\text{HCA}(y) = \frac{\text{Accuracy within } \mathcal{Y}_{\mathbf{x}_s} - \text{Chance Accuracy}}{1 - \text{Chance Accuracy}},
\label{eq:Revised_HCA}
\end{equation}
where \(\text{Accuracy within } \mathcal{Y}_{\mathbf{x}_s}\) is defined as:
  \begin{equation}
  \text{Accuracy within } \mathcal{Y}_{\mathbf{x}_s} = \frac{\arg\max_{y'\in \mathcal{Y}_{\mathbf{x}_s}} \text{CM}_{y',y}}{ N_y}.
  \end{equation}
\( \text{CM}_{y', y} \) is the count of predictions being \( y' \) for samples with true label \( y \), \( N_y \) is the total number of samples with true label \( y \), and \(\text{Chance Accuracy}\) is the accuracy that would be achieved by random guessing within the subset \( \mathcal{Y}_{\mathbf{x}_s} \), given by \(\text{Chance Accuracy} = \left|\mathcal{Y}_{\mathbf{x}_s}\right|^{-1}\), where \( |\mathcal{Y}_{\mathbf{x}_s}| \) is the number of labels in the subset \( \mathcal{Y}_{\mathbf{x}_s} \). This formulation of \(\text{Accuracy within } \mathcal{Y}_{\mathbf{x}_s}\) captures the maximum accuracy achieved by the most frequently predicted label within \( \mathcal{Y}_{\mathbf{x}_s} \) for samples with true label \( y \). By comparing this accuracy to the chance level, we assess whether the network's performance exceeds what would be expected by random guessing within the subset. 

Then, We compute the \textbf{Average Hierarchical Classification Accuracy (AHCA)} over all true labels:
\begin{equation}
\text{AHCA} = \frac{1}{|\mathcal{Y}|} \sum_{y \in \mathcal{Y}} \text{HCA}(y).
\label{eq:AHCA}
\end{equation}

\item \textbf{Scenario B.} In this scenario, the labels are assigned based on the simple features \( \mathbf{x}_s \), which are fully predictive of the true labels. The complex features \( \mathbf{x}_c \) introduce spurious correlations but should not influence the network's predictions due to simplicity bias. To confirm that the network relies solely on \( \mathbf{x}_s \), we define the \textbf{Prediction Consistency Score (PCS)} as:
\begin{equation}
\text{PCS} = \frac{1}{|\mathcal{Y}|} \sum_{y \in \mathcal{Y}} \frac{\text{CM}_{y,y}}{N_y}.
\end{equation}
\end{itemize}

A high AHCA indicates that the network consistently uses the complex features \( \mathbf{x}_c \) to make accurate predictions within each group defined by \( \mathbf{x}_s \). A high PCS indicates that the network's predictions are accurate and consistent with the true labels determined by \( \mathbf{x}_s \), suggesting that \( \mathbf{x}_c \) does not adversely affect the network's decision-making process. To make the experiment more interpretable, we present both measures (\text{AHCA} and \text{PCS}) for each scenario in the experiments. This approach ensures that the roles of complex features and simple features can not be interchanged, thereby accurately capturing the hierarchical simplicity bias in different experimental setups.

\section{Experiment}
\label{sec:exp}

\subsection{Experiment Setup}

\begin{figure}[htb]
    \centering
    \includegraphics[width=.35\textwidth]{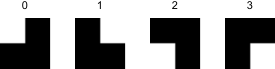}
    \caption{The patch data types.}
    \label{fig:patch}
\end{figure}

\paragraph{Building blocks.} We utilize three primary datasets to construct our synthetic datasets: \textbf{(1)} Patch: Four types of deterministic patches, each featuring a white corner with the remaining area in black (Figure \ref{fig:patch}). The patch data is deterministic. \textbf{(2)} MNIST \citep{lecun2010mnist}. \textbf{(3)} CIFAR-10 \citep{krizhevsky2009learning}. 

\begin{figure}[htbp]
    \centering
    \begin{minipage}{0.225\textwidth} 
        \centering
        \includegraphics[width=\linewidth]{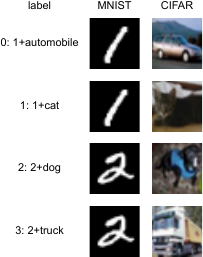}
    \end{minipage}
    \hfill 
    \begin{minipage}{0.225\textwidth} 
        \centering
        \includegraphics[width=\linewidth]{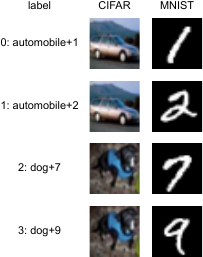}
    \end{minipage}
    \hfill
    \begin{minipage}{0.225\textwidth} 
        \centering
        \includegraphics[width=\linewidth]{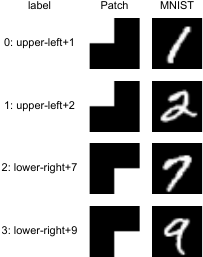}
    \end{minipage}
    \hfill
    \begin{minipage}{0.225\textwidth} 
        \centering
        \includegraphics[width=\linewidth]{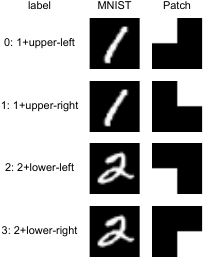}
    \end{minipage}
    \caption{Illustration of four datasets. From left to right: (a) MNIST-CIFAR, (b) CIFAR-MNIST, (c) Patch-MNIST, (d) MNIST-Patch.}
    \label{fig:datasets}
\end{figure}

\paragraph{Synthetic Datasets.} We create four training datasets by combining the building blocks using im-
balanced label coupling, as shown in Figure \ref{fig:datasets}. Details of the experimental setup, including data preprocessing and model configurations, are
provided in Appendix \ref{sec:expsetup}.

\subsection{Results and Discussion}

\begin{figure}[htb]
    \centering
    \includegraphics[width=.9\textwidth]{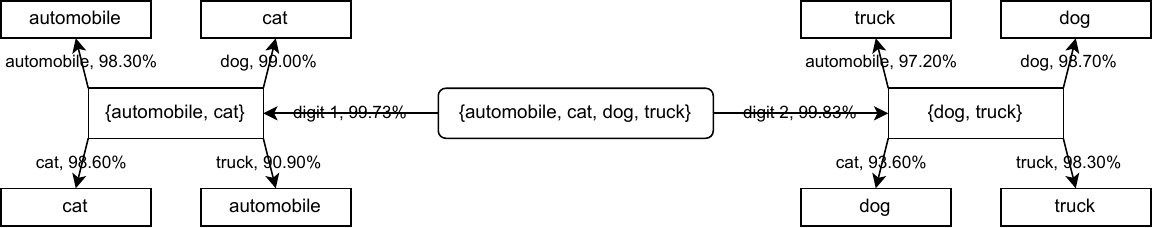}
    \caption{The inferred decision tree from the neural network trained on the MNIST-CIFAR dataset. The boxes contain the network's predictions, while the arrows indicate the path taken based on the input features. The percentages show the proportion of samples following each path. For example, among all samples with MNIST digit 1, approximately 99.73\% are predicted to be either automobile or cat. Within this group, 99.00\% of samples where the CIFAR-10 image is a dog are misclassified as cat.}
    \label{fig:2}
\end{figure}

In the MNIST-CIFAR dataset shown in Figure \ref{fig:datasets}(a), we combine two datasets: CIFAR-10 images and MNIST digits. Here, CIFAR-10 images (\(\mathbf{x}_c\)) represent the complex features (fine dataset), while MNIST digits (\(\mathbf{x}_s\)) represent the simple features (coarse dataset). Each MNIST digit is linked to a specific subset of CIFAR-10 labels; for example, the digit 1 corresponds to \(\mathcal{Y}_{\mathbf{x}_s} = \{\text{automobile}, \text{cat}\}\). As shown in Figure \ref{fig:2}, the trained neural network exhibits a hierarchical decision-making process akin to a decision tree. Initially, the network uses the simple MNIST digit (\(\mathbf{x}_s\)) to narrow down the possible labels to a subset. Then, within this subset, it uses the complex CIFAR-10 image (\(\mathbf{x}_c\)) for fine-grained classification.

Notably, when misclassifications occur, they are not random but instead exhibit patterns that reflect the retained predictive power of the complex features. For instance, many images of automobiles are misclassified as trucks, and cats are misclassified as dogs. These specific errors suggest that the network is leveraging semantic similarities among CIFAR-10 classes within the subsets determined by the MNIST digits.

To quantitatively assess the hierarchical simplicity bias, we calculate the Average Hierarchical Classification Accuracy (AHCA) and Prediction Consistency Score (PCS) for each dataset, as shown in Table \ref{fig:tables}(a). Recall that AHCA measures the network's ability to use complex features within groups defined by simple features, while PCS measures the reliance on simple features when they are fully predictive.

\begin{figure}[htbp]
    \centering
    \begin{minipage}{0.5\textwidth}
        \centering
        \begin{tabular}{ccc}
            \toprule
            Dataset     & AHCA  & PCS    \\ \midrule
            MNIST-CIFAR & 93.65 & 49.35  \\
            CIFAR-MNIST & 0.93  & 99.24  \\
            Patch-MNIST & 78.05 & 49.96  \\
            MNIST-Patch & 0.00  & 100.00 \\ \bottomrule
        \end{tabular}
    \end{minipage}%
    \begin{minipage}{0.5\textwidth}
        \centering
        \begin{tabular}{ccc}
            \toprule
            Model    & Standard & Semantic \\
            \midrule
            Spurious & 49.24    & 96.83    \\
            DFR      & 68.12    & 97.55    \\
            Baseline & 87.75    & 98.30    \\
            \bottomrule
        \end{tabular}
    \end{minipage}
    \caption{(a) Left: AHCA and PCS (\%) for four different datasets. (b) Right: Performance comparison of three models: Spurious (trained on the MNIST-CIFAR dataset), DFR (the Spurious model after DFR), and Baseline (trained on CIFAR-10 images).}
    \label{fig:tables}
\end{figure}

In scenarios where labels are based on complex features (e.g., \textit{MNIST-CIFAR} and \textit{Patch-MNIST}), the high AHCA values indicate that networks effectively utilize complex features for classification within the groups defined by simple features. Conversely, in scenarios where labels are based on simple features (e.g., \textit{CIFAR-MNIST} and \textit{MNIST-Patch}), the high PCS values and low AHCA values illustrate that networks predominantly depend on simple features and largely ignore complex features. More detailed experiment results are in Appendix \ref{sec:expdetail}.

Building on our observations of hierarchical simplicity bias, we investigate whether existing methods can mitigate the impact of spurious correlations in such hierarchical settings.

\subsection{Last-Layer Retraining Is Insufficient for Strong Spurious Correlations}
\label{sec:dfr}

\citet{kirichenko2022last} introduced Deep Feature Reweighting (DFR), a method aimed at improving model robustness against spurious correlations by retraining the last layer using data from the target distribution.  We apply DFR to our MNIST-CIFAR dataset (Figure \ref{fig:datasets}(a)) to assess its effectiveness in our hierarchical setting. In this dataset, classes 0 and 3 correspond to vehicles (automobile and truck), and classes 1 and 2 correspond to animals (dog and cat). We define a \emph{semantically correct prediction} as correctly classifying a vehicle as a vehicle or an animal as an animal, regardless of the specific class label. It allows us to evaluate whether the network retains an understanding of core features at a higher level of abstraction. The results are presented in Table \ref{fig:tables}(b).

Although DFR improves the standard accuracy from $49.24\%$ to $68.12\%$, it does not achieve the baseline accuracy of $87.75\%$, suggesting that DFR alone may not fully address the challenges posed by perfectly correlated spurious features. This limitation indicates that some core feature information may not be adequately captured in the final-layer representations, possibly due to loss of information in intermediate layers when spurious features are very strong.

Interestingly, the high semantic accuracies suggest that the neural network retains an understanding of core features at a higher level of abstraction. For the Spurious model, the semantic accuracy is $96.83\%$, which is very close to the baseline model's $98.30\%$. This indicates that despite the influence of spurious features on fine-grained class distinctions, the network is still capable of distinguishing between broader categories such as vehicles and animals. This observation underscores the hierarchical nature of the network's decision-making process, where it can process and retain information at multiple levels of complexity. It also highlights the importance of evaluating model performance across different levels of abstraction to gain deeper insights into neural network behavior beyond overall accuracy metrics.

\section{Conclusion and Future Work}

\paragraph{Conclusion.} We introduced \emph{imbalanced label coupling} to extend simplicity bias to multiple hierarchical levels, demonstrating that neural networks exhibit a hierarchical simplicity bias. Our experiments with synthetic datasets show that networks prioritize features based on ascending complexity correlated with labels, mirroring a decision tree's behavior. Moreover, we found that last-layer retraining is insufficient to recover core features when spurious correlations are perfect, indicating that core feature information may be lost in intermediate layers. Our study provides insights into how neural networks prioritize features of varying complexities, contributing to a deeper understanding of implicit bias and aiding the development of more robust machine learning systems.

\paragraph{Limitations and Future Work.} The extreme hierarchical relationships observed in our study may not be easily observable in practice, as our research is based on synthetic datasets designed to highlight certain behaviors in a controlled setting. Future work could extend these findings to real-world situations, establish a theoretical framework for hierarchical simplicity bias, and propose strategies to mitigate adverse effects of this bias.


\bibliography{mybib}
\newpage
\appendix

\section{Related Work}
\label{sec:relate}

\paragraph{Simplicity bias.} 
\citet{kalimeris2019sgd} show that stochastic gradient descent learns functions of increasing complexity. \citet{soudry2018implicit,gunasekar2018implicit,nacson2019convergence,lyu2021gradient,kunin2022asymmetric} show that gradient descent leads to max-margin classifiers. \citet{gunasekar2017implicit,arora2019implicit,huh2021low} show that deep networks are inductively biased to find low-rank solutions. \citet{valle2018deep} show that the parameter-function map is biased towards simple functions with algorithmic information theory. \citet{pezeshki2021gradient} propose gradient starvation. \citet{morwani2023simplicity} rigorously study simplicity bias in one hidden layer neural networks. \citet{cao2019towards,rahaman2019spectral} study the spectral bias of deep neural networks.

Taking a step further, our study reveals a more nuanced perspective. Complex features are not entirely ignored but can retain their predictive power. Our findings do not contradict prior research where simple features have complete predictive power. Rather, in our experiments, simple features serve as predictive indicators only for the specific subgroups trained with those features. Thus, our work extends earlier research: where extreme simplicity bias is similar to a single-level decision tree, our approach resembles a multi-level decision tree.

\paragraph{Spurious correlations.}

Neural networks exhibit bias towards spurious features, which are only associated with the task label but are not causally related \citep{geirhos2020shortcut}. Such biases have been found in real-world image datasets \citep{singla2021salient,singla2021understanding}. Such features include texture \citep{geirhos2018imagenet}, poses \citep{alcorn2019strike}, and background \citep{geirhos2018imagenet,moayeri2022comprehensive}. Overparametrization's negative impact on model performance in the context of these spurious correlations is investigated by \citet{sagawa2020investigation}. Further, studies have been conducted to understand the influencing factors of feature representations \citep{hermann2020shapes} and to identify likely shortcut cues \citep{scimeca2021shortcut}. Mitigation methods include distributionally robust optimization (DRO) targeting worst-group loss instead of average loss \citep{ben2013robust,hu2018does,sagawa2019distributionally,zhang2020coping}, invariant learning \citep{arjovsky2019invariant,koyama2020out,creager2021environment,yao2022improving}, and weighting \citep{nam2020learning,liu2021just,zhang2022correct}.

In closely related works, \citet{kirichenko2022last,rosenfeld2022domain} demonstrate that despite spurious correlations, neural networks capture core features that can be recovered by retraining the final layer with target distribution data. Our work, on the other hand, investigates the confusion matrix, highlighting the network's inherent reliance on true object semantics in the presence of spurious features, without training a feature representation decoder. We also show that last-layer retraining is insufficient to fully capture the network's ability to utilize
core features. Similarly, \citet{ye2023freeze} provide a theoretical analysis of last-layer retraining for two-layer networks, proving that core features are only learned well when their associated non-realizable noise is small.

\section{Experiment Setup\label{sec:expsetup}}
\paragraph{Training set.} The training set is created through \emph{imbalanced label coupling}. We choose two different datasets: one with coarse labels and another with fine labels. For each class from the coarse dataset, we concatenate it with multiple classes from the fine dataset to create the training examples. We assign labels to these examples based on the fine dataset. In other words, the fine dataset has full predictive power, while the coarse dataset does not. Across all experiments, every class comprises $5000$ training examples, and the total number of classes is the number of classes chosen in the fine dataset. This construction can be naturally extended to three or more datasets.

Our patch dataset includes four types of patch data, each sized at $32\times 32$ and $1$ channel. The MNIST dataset's images are subjected to zero padding, extending them from their original dimensions of $28\times 28$ to $32\times 32$. 

For instance, we form the MNIST-CIFAR dataset by concatenating the CIFAR-10 image with dimensions $3\times32\times32$ and the expanded MNIST digit with dimensions $1\times32\times32$ to generate a new image with dimensions $4\times32\times32$.

\paragraph{Test set.} The test set is created by concatenating the image channels from all selected classes in each dataset, without any coupling constraints. Each potential combination results in 1000 test samples. Hence, the overall test sample count equals 1000 multiplied by the product of the selected class counts in each dataset.

\paragraph{Network architecture and training configuration.} We use the ResNet-18 architecture \citep{he2016deep}, only making slight adjustments to align the channel of the initial convolutional filter with the input and output dimensions to match the class count. We train the model for 150 epochs using the SGD optimizer with a momentum of 0.9, employing a batch size of 128. The initial learning rate is $0.1$, and is reduced by a factor of $10$ at epochs $50$ and $100$. We apply a weight decay of $0.0005$. The loss function used for training is the cross-entropy loss. For more experiments on the multilayer perceptron (MLP), see Appendix \ref{sec:appendix}.

\section{Detailed Experiment Results \label{sec:expdetail}}

This section presents the comprehensive results of the experiments outlined in Section \ref{sec:exp}. We include the confusion matrices for all experimental setups, which are utilized to calculate the Average Hierarchical Classification Accuracy (AHCA) and Prediction Consistency Score (PCS). These metrics provide a quantitative evaluation of the neural networks' hierarchical simplicity bias, illustrating how predictions align with feature complexity and the extent to which simple and complex features influence classification outcomes. 

\begin{figure}[htbp]
    \centering
    \begin{minipage}{0.25\textwidth} 
        \centering
        \includegraphics[width=\linewidth]{mnist2cifar4_train-eps-converted-to.pdf}
        \label{fig:1_1}
    \end{minipage}
    \hfill 
    \begin{minipage}{0.7\textwidth} 
        \centering
        \includegraphics[width=\linewidth]{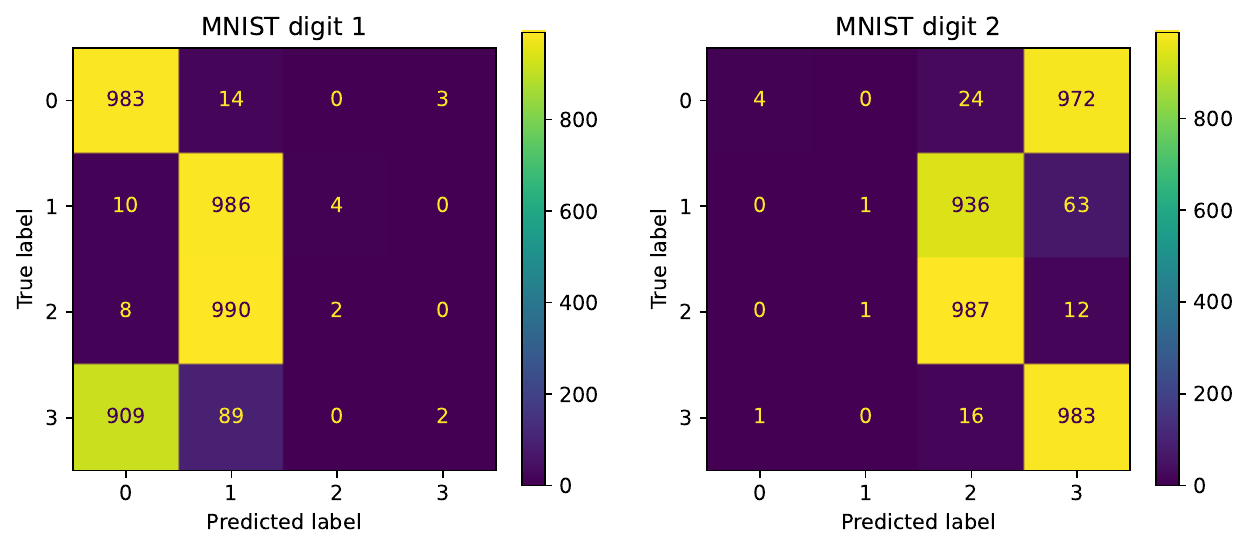}
        \label{fig:1_2}
    \end{minipage}
    \caption{The MNIST-CIFAR training set and test results. \textbf{(a)} Left: within the training set, digit 1 is concatenated with both automobile and cat images, while digit 2 is concatenated with both dog and truck images. \textbf{(b)} Right: during testing, most digit 1 samples are classified 0 and 1, while most digit 2 samples are classified 2 and 3, regardless of the corresponding CIFAR-10 image class.}
    \label{fig:mnistcifar}
\end{figure}

\begin{figure}[htbp]
    \centering
    \begin{minipage}{0.25\textwidth} 
        \centering
        \includegraphics[width=\linewidth]{cifar2mnist4_train-eps-converted-to.pdf}
    \end{minipage}
    \hfill 
    \begin{minipage}{0.7\textwidth} 
        \centering
        \includegraphics[width=\linewidth]{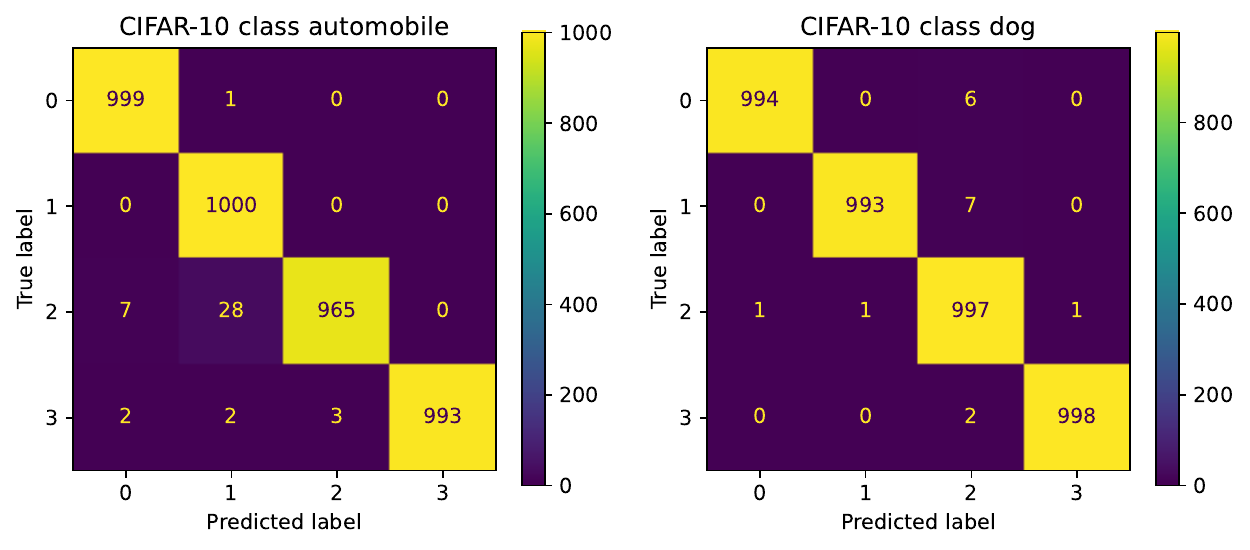}
    \end{minipage}
    \caption{The CIFAR-MNIST training set and test results. \textbf{(a)} Left: the training set includes automobile images paired with 1 and 2 digits, along with cat images paired with 7 and 9 digits. \textbf{(b)} Right: during testing, predictions rely mostly on MNIST and exhibit high accuracy.}
    \label{fig:cifarmnist}
\end{figure}

\paragraph{MNIST-CIFAR hierarchy.} We use MNIST as the coarse data and CIFAR-10 as the fine data in the MNIST-CIFAR training set, shown in Figure \ref{fig:mnistcifar}. Conversely, the CIFAR-MNIST training set is shown in Figure \ref{fig:cifarmnist}. The neural network consistently gives priority to MNIST regardless of its predictive power. The CIFAR-10 part only comes into play when MNIST's predictive power is insufficient. Hence, MNIST exhibits a strictly simpler complexity than CIFAR-10.

\begin{figure}[htbp]
    \centering
    \begin{minipage}{0.25\textwidth} 
        \centering
        \includegraphics[width=\linewidth]{patch2mnist4_train-eps-converted-to.pdf}
    \end{minipage}
    \hfill 
    \begin{minipage}{0.7\textwidth} 
        \centering
        \includegraphics[width=\linewidth]{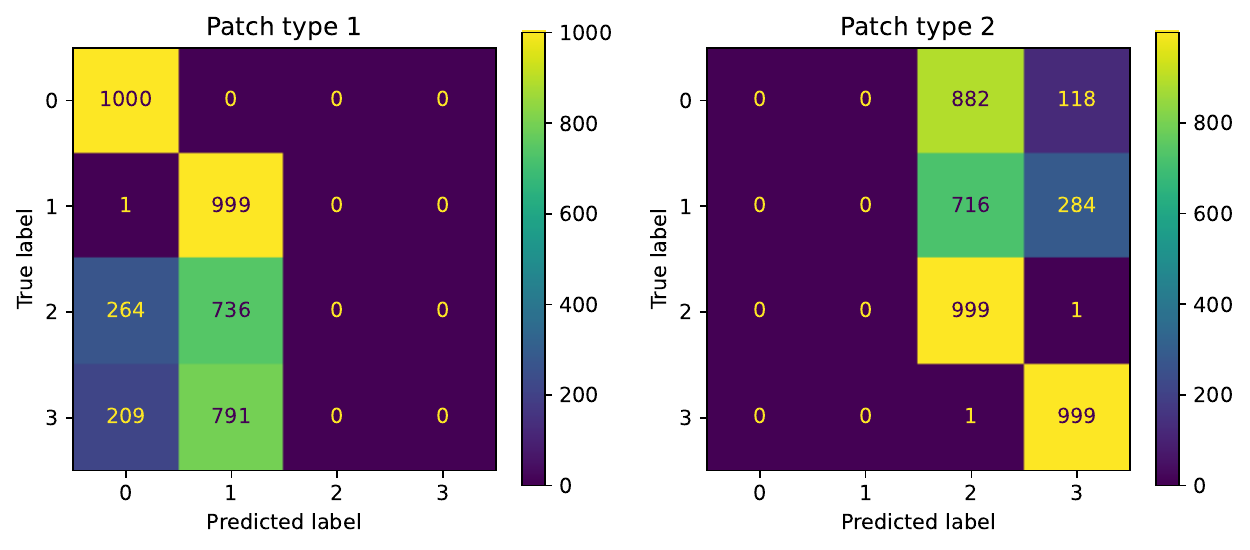}
    \end{minipage}
    \caption{The Patch-MNIST training set and test results. \textbf{(a)} Left: the upper-left patch is paired with 1 and 2 digits, along with the lower-right patch paired with 7 and 9 digits. \textbf{(b)} Right: in testing, all predictions prioritize patch data without any exceptions.}
    \label{fig:patchmnist}
\end{figure}

\begin{figure}[htbp]
    \centering
    \begin{minipage}{0.25\textwidth} 
        \centering
        \includegraphics[width=\linewidth]{mnist2patch4_train-eps-converted-to.pdf}
    \end{minipage}
    \hfill 
    \begin{minipage}{0.7\textwidth} 
        \centering
        \includegraphics[width=\linewidth]{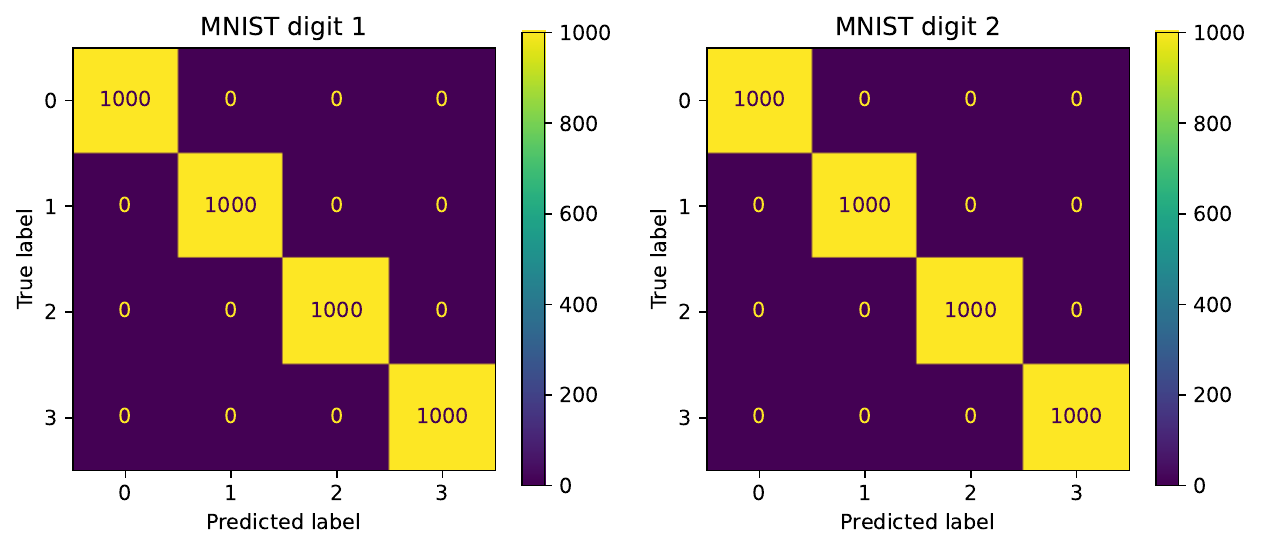}
    \end{minipage}
    \caption{The MNIST-Patch training set and test results. \textbf{(a)} Left: digit 1 is paired with upper-left and upper-right patches, and digit 2 is paired with lower-left and lower-right patches. \textbf{(b)} Right: the test accuracy in achieves $100\%$ using solely patch data.}
    \label{fig:mnistpatch}
\end{figure}

\paragraph{Patch-MNIST Hierarchy.} Similarly, we construct the training set of Patch-MNIST and MNIST-Patch. Figure \ref{fig:patchmnist} and Figure \ref{fig:mnistpatch} show the training set and test results. These results highlight the hierarchy that patch data is inherently less complex than MNIST. Interestingly, predictions are not completely random with spurious patch data in Figure \ref{fig:patchmnist}. For instance, when comparing digits 1 and 2, digit 2 has the closest visual similarity to digits 7 and 9 (and vice versa for digit 7). As a result, a large portion of digits 7 and 9 are categorized under digit 2 rather than digit 1.

\begin{figure}[htbp]
    \centering
    \includegraphics[width=\textwidth]{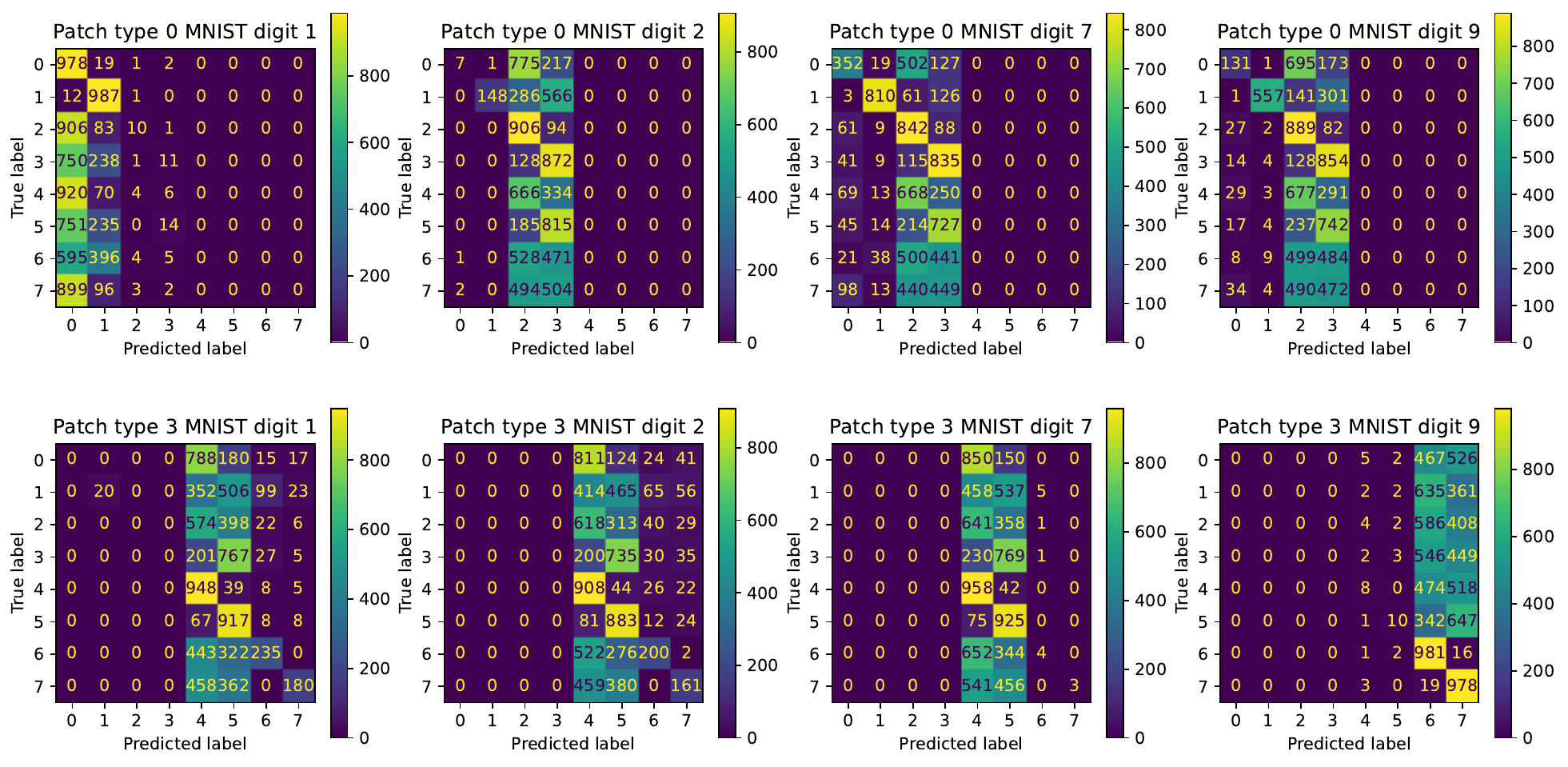}
    \caption{The confusion matrices of the network trained on the Patch-MNIST-CIFAR dataset.}
    \label{fig:patch2mnist4cifar8}
\end{figure}

\paragraph{Patch-MNIST-CIFAR hierarchy.} We construct a dataset named Patch-MNIST-CIFAR, containing all aforementioned building blocks, revealing a three-level hierarchy. The upper-left patch is coupled with digits 1 and 2, while the lower-right patch is coupled with digits 5 and 9. These digits (1, 2, 5, 9) are then coupled with CIFAR-10 classes (0, 1), (2, 3), (4, 5), and (6, 7) respectively, resulting in an 8-class dataset. The results are shown in Figure \ref{fig:patch2mnist4cifar8}. The neural network demonstrates a distinct hierarchy in its predictions based on increasing feature complexity: first patch data, then MNIST digits, and finally CIFAR-10 images.

\begin{figure}[htbp]
    \centering
    \includegraphics[width=0.7\linewidth]{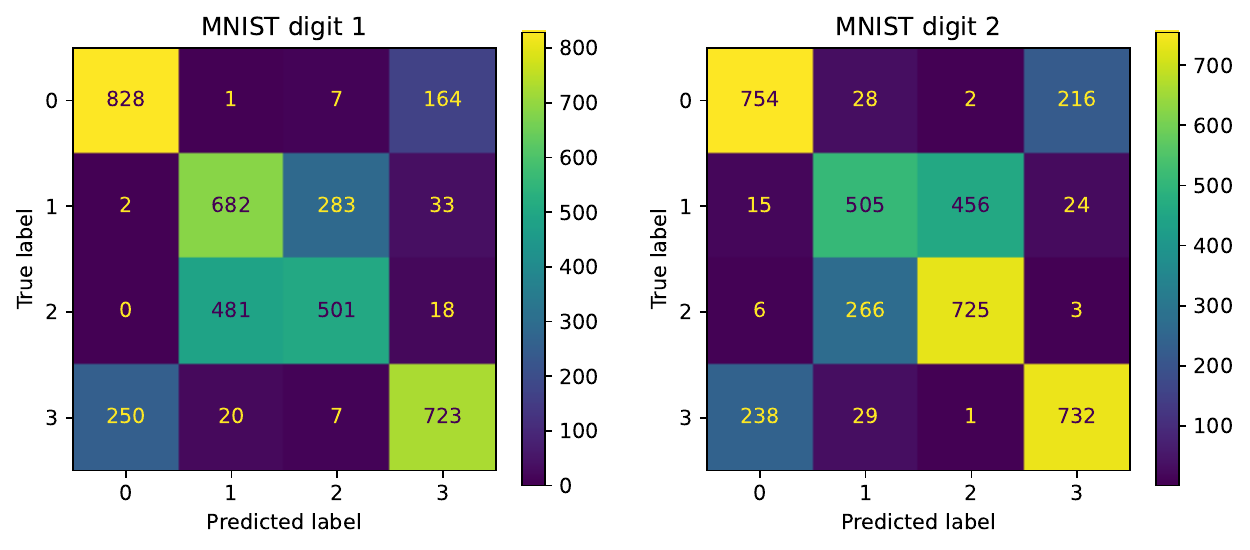}
    \label{fig:cifarmnistdfr}
    \caption{Confusion matrices following DFR of the neural network trained on the MNIST-CIFAR dataset in Figure \ref{fig:datasets}(a).}
\end{figure}

\section{Background and Corruption Hierarchy}
\label{sec:limit}

It has been demonstrated that neural networks exhibit a bias for texture and background \citep{geirhos2018imagenet,moayeri2022comprehensive}. In this section, we present similar evidence of decision-tree-like behavior in scenarios involving different background colors and image corruptions.

\subsection{Background Hierarchy in Half-Inverted MNIST}
\label{sec:mnist}

\begin{figure}[htbp]
    \centering
    \begin{minipage}{0.2\textwidth} 
        \centering
        \includegraphics[width=.7\linewidth]{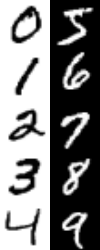}
        \caption{The training set.}
    \end{minipage}
    \hfill 
    \begin{minipage}{0.77\textwidth} 
        \centering
        \includegraphics[width=\linewidth]{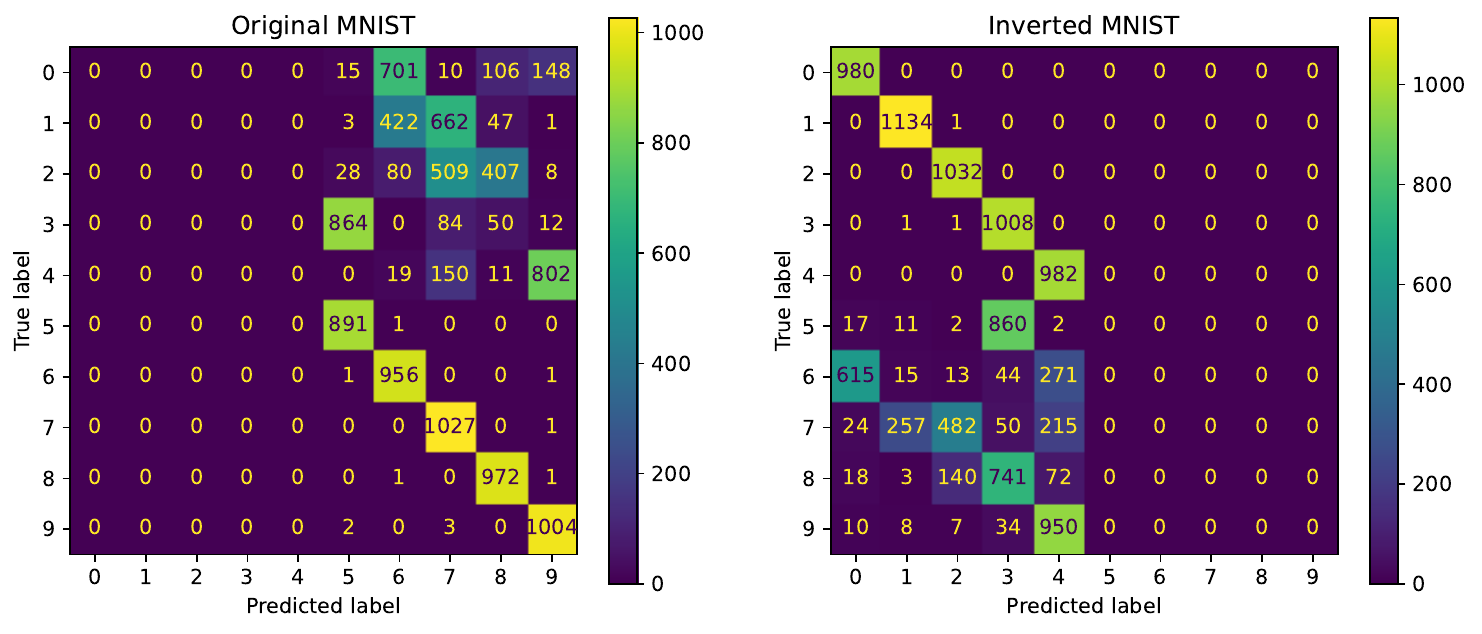}
        \caption{The confusion matrices on the original and color-inverted test set.}
    \end{minipage}
    \caption{The half-inverted MNIST dataset and test results. \textbf{(a)} Color inversion applied to digits 0-4; digits 5-9 remain unchanged. \textbf{(b)} During testing, the neural network demonstrates a consistent preference for spurious background colors, with digits of similar shapes exhibiting mutual misclassification.}
    \label{fig:mnistinverted}
\end{figure}

During training, we inverted the colors of MNIST digits 0-4 while leaving digits 5-9 unchanged. We tested the model using both the original and color-inverted test sets, with the results shown in Figure \ref{fig:mnistinverted}. The neural network consistently categorized white-background samples as 0-4 and black-background samples as 5-9, even though MNIST digits are nearly linearly separable. Unlike the ColorMNIST dataset in \citet{zhang2022correct} that uses five different background colors, our study uses just two (and also inverts the color of the digits themselves), which helps to better visualize the hierarchical decision-making process: most occurrences of digit 4 are wrongly classified as 9, and most of digit 3 as 5, and vice versa.

\subsection{Corruption Hierarchy in Corrupted CIFAR-10}
\label{sec:cifar}

\begin{figure}[htbp]
    \centering
    \includegraphics[width=\textwidth]{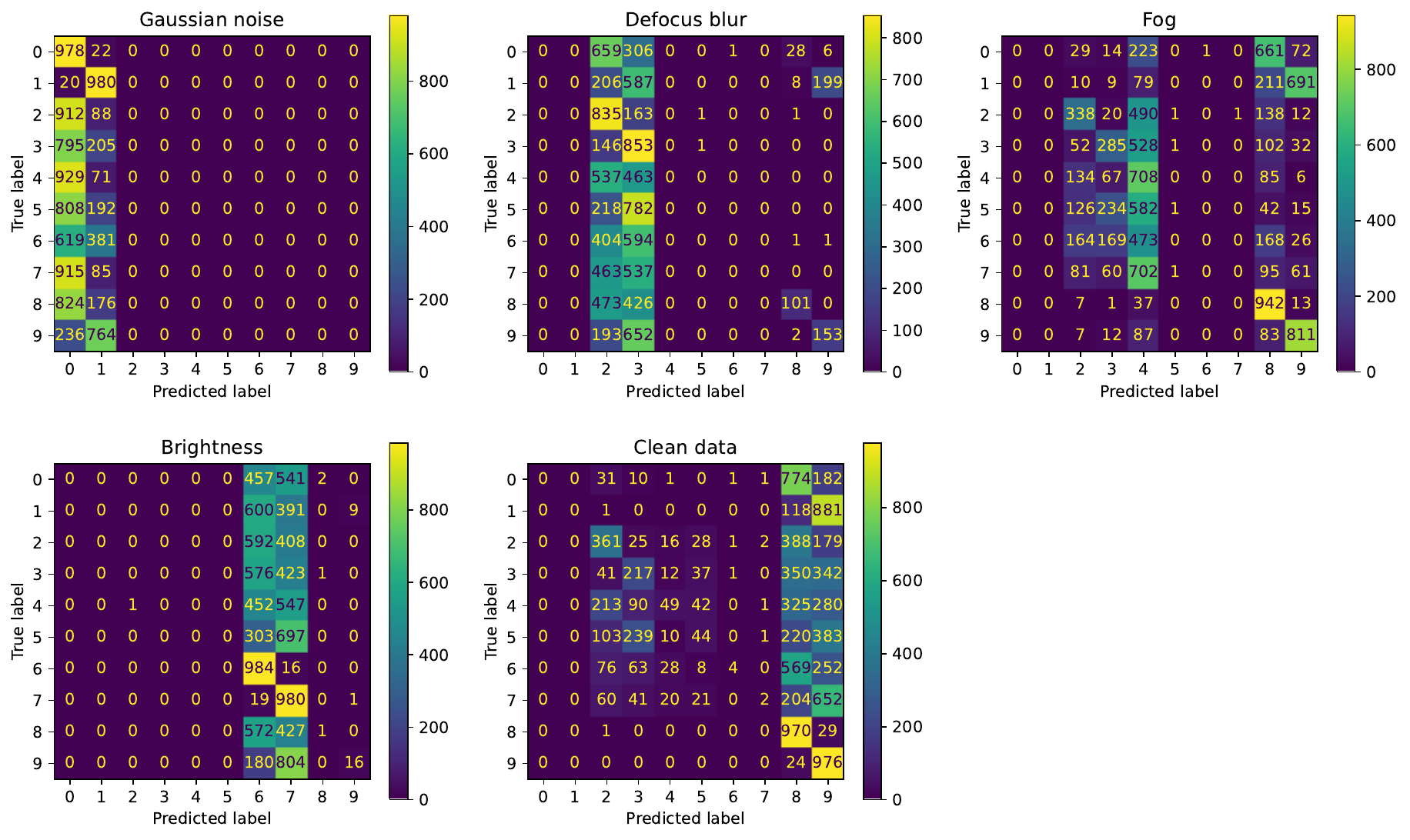}
    \caption{Test results from the neural network trained on the corrupted CIFAR-10 dataset demonstrate that most corruptions heavily influence classification, often leading to predictions falling within the two classes on which they were trained.}
    \label{fig:cifarcorruption}
\end{figure}

We subject the CIFAR-10 training set to four common types of corruptions \citep{hendrycks2019benchmarking}: Gaussian noise, Defocus blur, Fog, and Brightness, all with a severity level of $3$. For detailed implementations, see \citet{hendrycks2019benchmarking}. Each type of corruption is applied to specific CIFAR-10 classes: (0, 1), (2, 3), (4, 5), and (6, 7), while classes (8, 9) remain uncorrupted. The test set includes the original test set with $10,000$ samples, as well as the four corrupted versions of the test set, yielding a combined dataset of $50,000$ samples. The results are in Figure \ref{fig:cifarcorruption}. The hierarchical behavior is not easily visualizable under this setting as there are a total of five groups subjected to different corruptions. Nevertheless, the classification results are not uniformly randomly distributed within each group, which may provide evidence of the hierarchical classification process.

\section{Additional Experiments on MLP\label{sec:appendix}}
To ensure consistency of results across different architectures, we repeated all experiments, except for those in Subsection \ref{sec:dfr}, using a multi-layer perceptron (MLP). The MLP has 10 hidden layers, each containing a linear layer with a width of 1024, followed by batch normalization and ReLU activation. The final layer is linear. Other training setups are the same as those in Appendix \ref{sec:expsetup}. From the results below, we observe the hierarchical decision-making process in MLP, although it is not as pronounced as in the ResNet architecture for the corrupted CIFAR-10 dataset.

\begin{figure}[htbp]
    \centering
    \includegraphics[width=.7\linewidth]{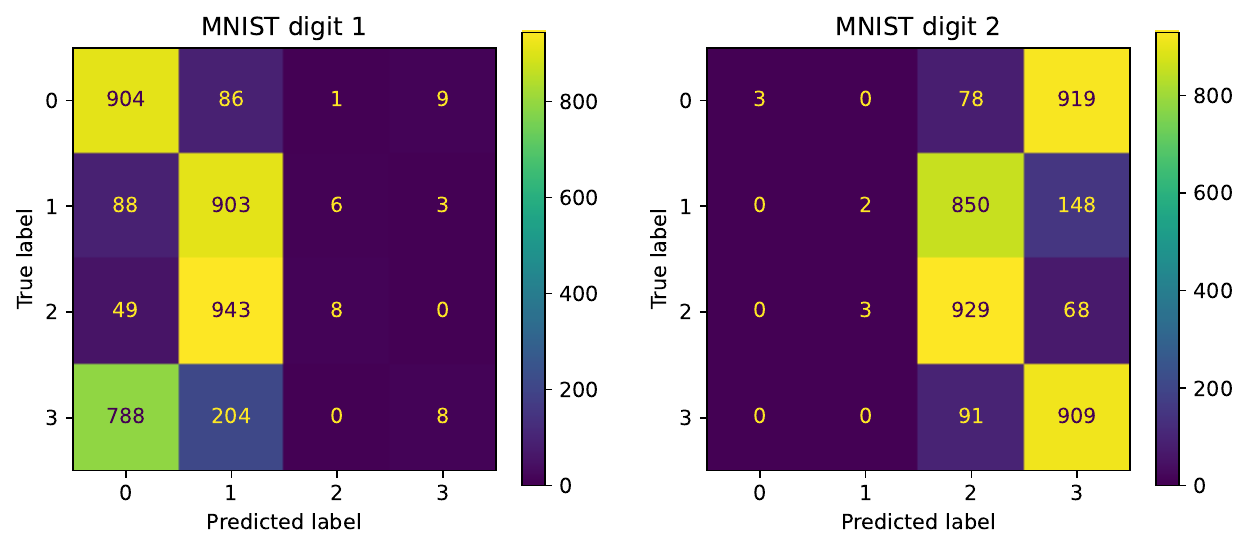}
    \caption{The confusion matrices for the MLP trained on the MNIST-CIFAR dataset.}
\end{figure}

\begin{figure}[htbp]
    \centering
    \includegraphics[width=.7\linewidth]{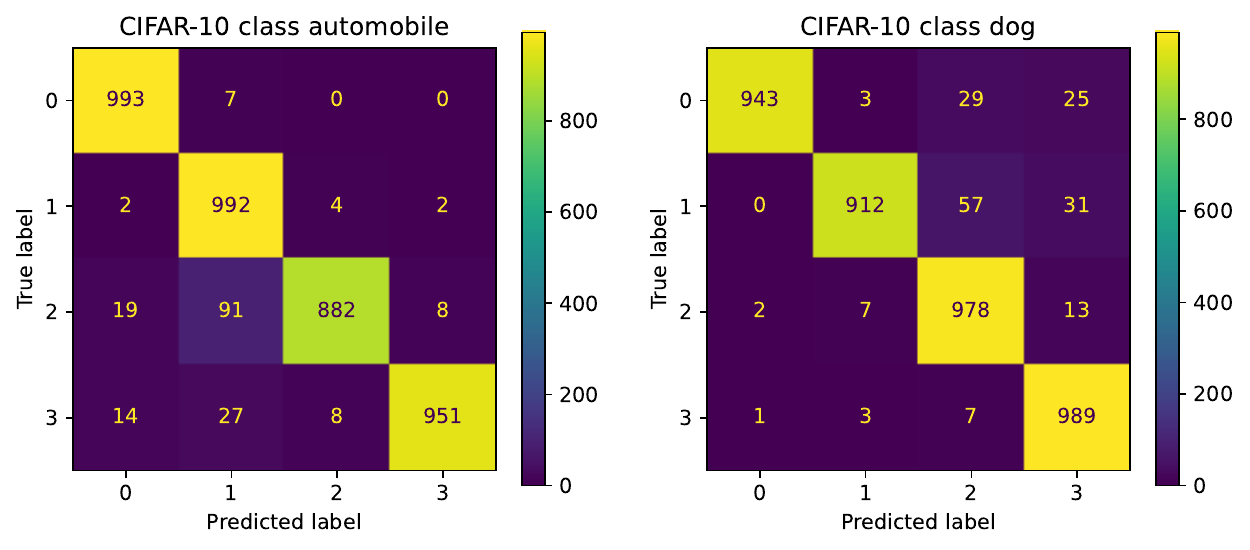}
    \caption{The confusion matrices for the MLP trained on the CIFAR-MNIST dataset.}
\end{figure}

\begin{figure}[htbp]
    \centering
    \includegraphics[width=.7\linewidth]{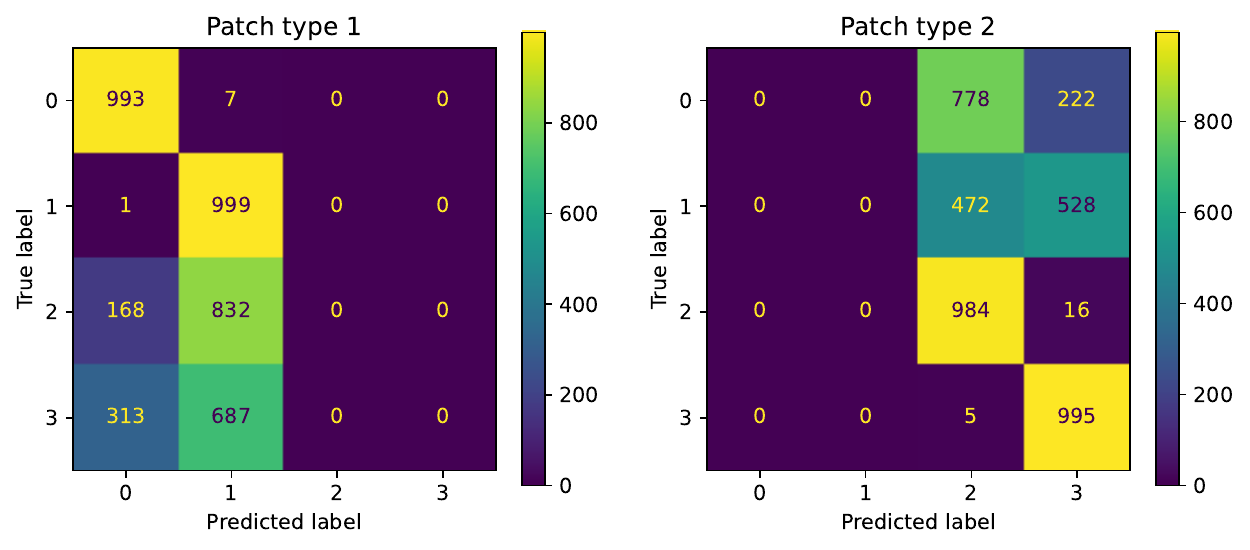}
    \caption{The confusion matrices for the MLP trained on the Patch-MNIST dataset.}
\end{figure}

\begin{figure}[htbp]
    \centering
    \includegraphics[width=.7\linewidth]{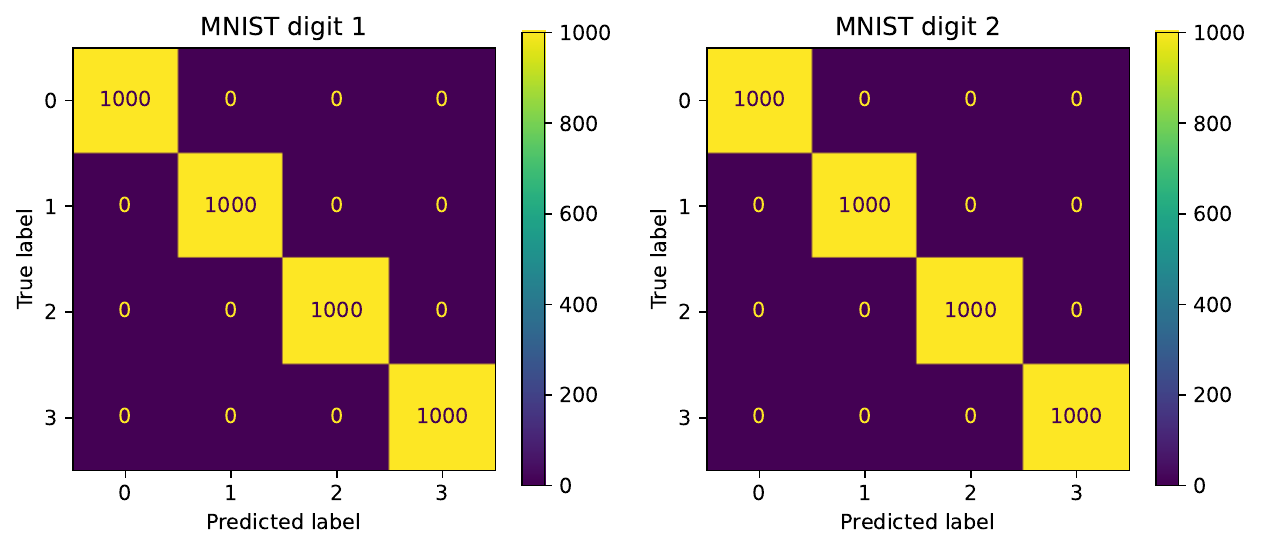}
    \caption{The confusion matrices for the MLP trained on the MNIST-Patch dataset.}
\end{figure}

\begin{figure}[htbp]
    \centering
    \includegraphics[width=\linewidth]{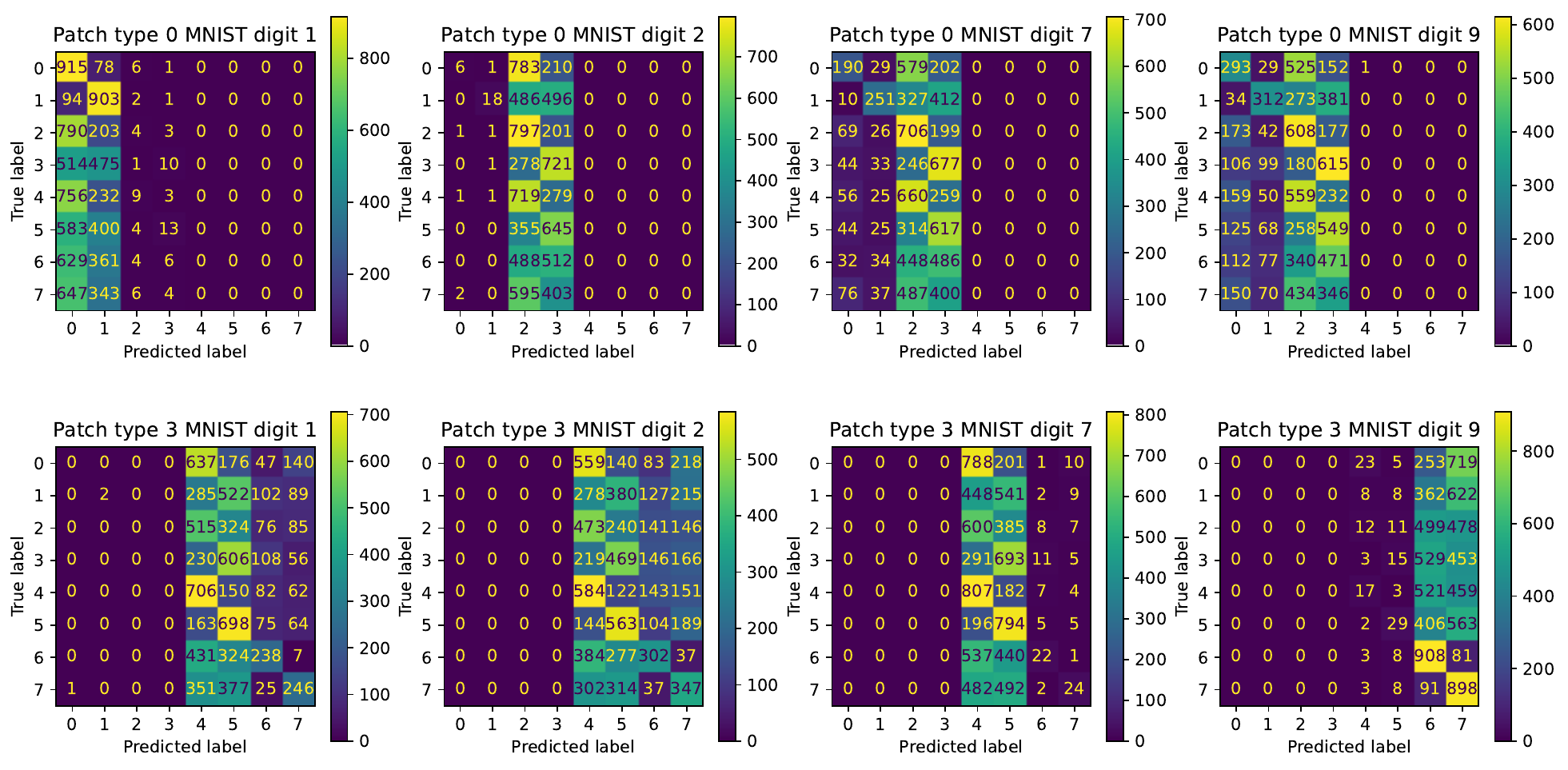}
    \caption{The confusion matrices for the MLP trained on the Patch-MNIST-CIFAR dataset.}
\end{figure}

\begin{figure}[htbp]
    \centering
    \includegraphics[width=\linewidth]{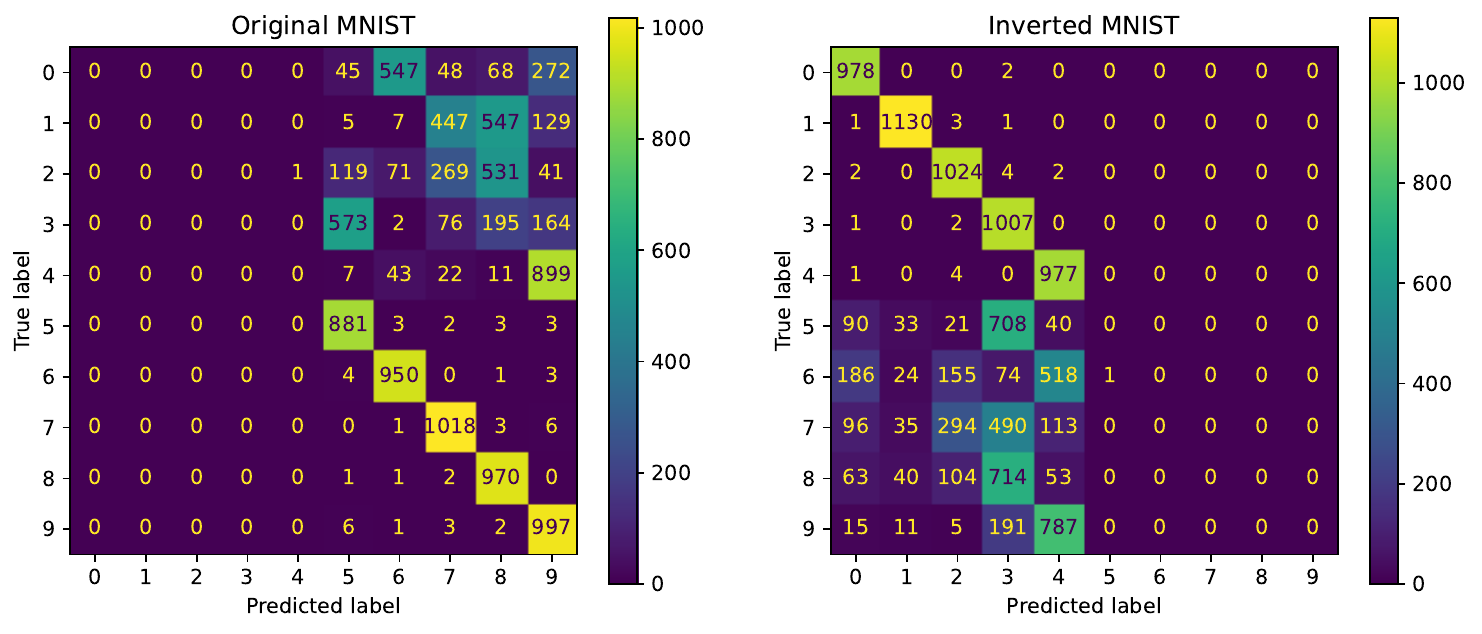}
    \caption{The confusion matrices for the MLP trained on the half-inverted MNIST dataset.}
\end{figure}

\begin{figure}[htbp]
    \centering
    \includegraphics[width=\linewidth]{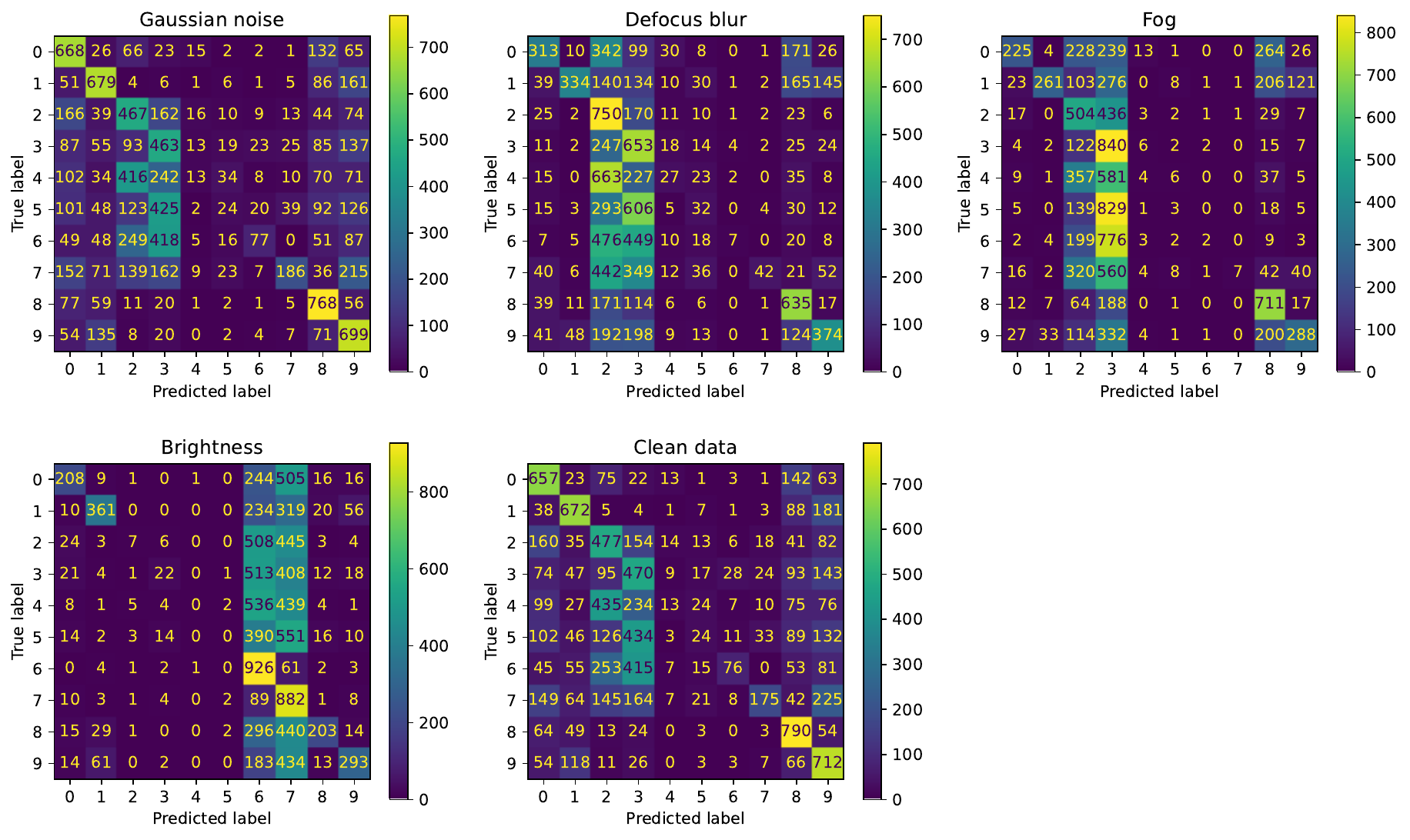}
    \caption{The confusion matrices for the MLP trained on the corrupted CIFAR-10 dataset.}
\end{figure}

\end{document}